\documentclass{article}

\usepackage{times}
\usepackage{graphicx} 
\usepackage{subfigure} 

\usepackage{natbib}

\usepackage{algorithm}
\usepackage{algorithmic}
\usepackage{amsmath}
\usepackage{amssymb}
\usepackage{booktabs}
\usepackage{multirow}
\usepackage{tabularx}
\usepackage{url}
\usepackage{tablefootnote}
\usepackage{array}
\newcolumntype{R}{>{\raggedleft\arraybackslash}p{.23cm}}

\usepackage{hyperref}

\usepackage[accepted]{icml2017}

\icmltitlerunning{Linguistic Knowledge as Memory for RNNs}

\begin{document} 

\twocolumn[
\icmltitle{Linguistic Knowledge as Memory for Recurrent Neural Networks} 


\icmlsetsymbol{equal}{*}

\begin{icmlauthorlist}
 \icmlauthor{Bhuwan Dhingra}{cmu}
 \icmlauthor{Zhilin Yang}{cmu}
 \icmlauthor{William W. Cohen}{cmu}
 \icmlauthor{Ruslan Salakhutdinov}{cmu}
\end{icmlauthorlist}

\icmlaffiliation{cmu}{School of Computer Science, Carnegie Mellon University, Pittsburgh, USA}

\icmlcorrespondingauthor{Bhuwan Dhingra}{bdhingra@cs.cmu.edu}

\icmlkeywords{boring formatting information, machine learning, ICML}

\vskip 0.3in
]



\printAffiliationsAndNotice{}  

\begin{abstract} 
    Training recurrent neural networks to model long term dependencies is difficult.
    Hence, we propose to use external linguistic knowledge as an explicit signal to inform the model which memories it 
    should utilize. Specifically, external knowledge is used to augment a sequence with typed edges between arbitrarily distant elements, 
    and the resulting graph is decomposed into directed acyclic subgraphs.
    We introduce a model that encodes such graphs as explicit memory in recurrent neural networks, and use it to model coreference relations in text.
    We apply our model to several text comprehension tasks and achieve new state-of-the-art results on all considered benchmarks, including CNN, bAbi, and LAMBADA.
    On the bAbi QA tasks, our model solves $15$ out of the $20$ tasks with only $1000$ training examples per task. Analysis of the learned representations
    further demonstrates the ability of our model to encode fine-grained entity information across a document.
\end{abstract} 

\section{Introduction}
Sequential data appears in many real world applications involving natural language, videos, speech and financial markets.
Predictions involving such data require accurate modeling of dependencies between elements of the sequence which may be arbitrarily far apart.
Deep learning offers the promise of extracting these dependencies in a purely data driven manner, with Recurrent Neural Networks (RNNs) being the
architecture of choice. RNNs show excellent performance when the dependencies of interest range short spans of the sequence,
however they can be notoriously hard to train to discover longer range dependencies \citep{koutnik2014clockwork,bengio1994learning}.

\citet{hochreiter1997long} introduced Long Short Term Memory (LSTM) networks which use a special unit called the Constant Error Carousel (CEC) to alleviate this problem.
The CEC has a memory cell with a constant linear connection to itself which allows gradients to flow over long distances. 
\citet{cho2014learning} introduced a simplified version of LSTMs called Gated Recurrent Units (GRU) with has one less gate, and consequently fewer parameters. Both LSTMs and GRUs have been hugely popular for modeling sequence data \citep{sutskever2014sequence,kiros2015skip,oord2016pixel}.

Despite these extensions, empirical studies have shown that it is still difficult to train RNNs with long-range dependencies (see for example, \citep{cho2014properties}). One suggested explanation for this is that the network must propagate all the information in a single fixed-size vector, which may be infeasible. This led to the introduction of the \textit{attention mechanism} \citep{bahdanau2014neural} which adapts the sequence model with a more explicit form of long term memory. At each time step $t$, the model can perform a ``soft''-lookup over all previous outputs through a weighted average $\sum_{i=1}^{t-1} \alpha_i h_i$. The weights $\alpha_i$ are the outputs of another network whose parameters are learned from data. Augmenting sequence models with attention has lead to significant improvements in various language modeling domains \citep{hermann2015teaching}. 
Other architectures, such as Memory Networks \citep{weston2014memory}, further build on this idea by introducing a memory module for the soft-lookup operation, and
a number of models allow the RNN to hold 
differentiable ``memories'' of past elements to
discover long range correlations \citep{graves2014neural}. 
However, \citet{daniluk2017frustratingly} showed that even memory-augmented neural models do not look beyond the immediately preceding time steps.
Clearly, training RNNs to discover long range dependencies without an explicit signal is challenging. 

\begin{figure*}[t!]
    \centering
\includegraphics[width=0.8\linewidth]{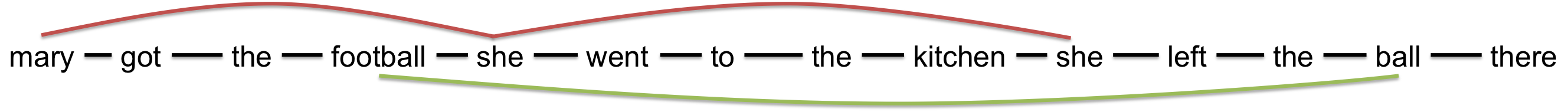}
\vspace{-0.3cm}
\caption{A sequence of text augmented with extra links. Red edges denote coreference relations, and green edges denote hypo/hyper-nymy. Black edges denote usual sequential links.}
\label{fig:dagexample}
\vspace{-0.3cm}
\end{figure*}

In this paper we do not attempt to solve this problem. Instead we argue that in many applications, information about long-term dependencies may be readily available in the form of symbolic knowledge. As an example, consider the sequence of text
shown in Figure~\ref{fig:dagexample}. Standard preprocessing tools can be used to extract relations such as coreference and hypernymy between pairs
of tokens, which can be added as extra edges in addition to the sequential links between elements. 
We argue that these extra signals can be used to provide the RNN model with locations of an \textit{explicit memory} of distant elements
when computing the representation of the current element. The content of a memory is the representation of the linked element, and edge labels
can distinguish different types of memories. In this manner symbolic knowledge can guide the propagation of information through a recurrent network.

Technically, incorporating these ``skip connections'' into the sequence converts it into a graph with cycles. Graph based neural networks \cite{scarselli2009graph} can be used to 
handle such data, but they are computationally expensive when the number of nodes in the graph is large. Instead, we utilize the order inherent in the
the unaugmented sequence to decompose the graph into two Directed Acyclic Graphs (DAGs) with a topological ordering. We introduce the
Memory as Acyclic Graph Encoding RNN (MAGE-RNN) framework
to compute the representation of such graphs while touching every node only once, and implement a GRU version of it called MAGE-GRU. MAGE-RNN learns separate
representations for propagation along each edge type, which leads to superior performance empirically. In cases where there is at most a single 
incoming edge of a particular type at a node, it reduces to a memory augmented regular RNN whose memory access is determined by a symbolic signal.

We use MAGE-RNN to model coreference relations for text comprehension tasks, where answers to a query have to be extracted from a context document. 
Tokens in a document are connected by a coreference relation if they refer to the same underlying entity. Identifying such relations is important
for developing an understanding of the document, and hence we augment RNN architectures for 
text comprehension with an explicit memory of coreferent mentions. MAGE-GRU leads to a consistent improvement over the vanilla GRU, as well as a baseline
where the coreference information is added as input features to the model. By further replacing GRU units in existing reading comprehension models with MAGE-GRUs
we achieve state-of-the-art performance on three well studied benchmarks -- the bAbi QA tasks, the LAMBADA dataset, and the CNN dataset. An analysis of
the learned representations by the model also show its effectiveness in encoding fine-grained information about the entities in a document.

\section{Related Work}

Augmenting the sequence with these extra links converts it from a chain to a more general graph structure. 
Models such as the Graph Neural Networks \citep{scarselli2009graph} and Gated Graph Sequence Neural Networks \citep{li2015gated}
can be used to handle such data. The basic idea in these architectures is to update the representation of every single node
at every time-step, based on the incoming representations of their neighbours. Depending on the optimization procedure, the 
updates are either performed till the representations converge, or for a fixed number of time-steps. The resulting complexity is
$O(N T)$, where $N$ is the number of nodes, and $T$ is the number of time-steps. To fully propagate information in the graph, $T$ should be at least the width of the graph. Though in practice it is possible to obtain an approximation with a smaller $T$, training graph-based neural networks is computationally expensive compared to RNNs, which have complexity only $O(N)$.

Trees are another commonly encountered graph structure, and \citet{tai2015improved} proposed Tree-Structured LSTMs for handling such data. However, that work focused solely on dependency parses of textual data and ignored its inherent sequential ordering, whereas here we argue for a more general approach which can incorporate many types of edges between tokens, including sequential ones. The resulting MAGE-RNN formulation can be viewed as an extension of Tree-Structured LSTMs.

\citet{shuai2016dag} proposed a similar idea to ours, which employs RNNs on DAGs of image pixels. However, their model focuses on images and does not handle typed edges. In contrast, our work provides a novel perspective on incorporating symbolic knowledge as links to sequential data. Moreover, in terms of model architectures, our model explicitly handles typed edges by using separate parameter matrices and split hidden states, which are key to improved performance.


The importance of reference resolution for reading comprehension was previously studied in \citet{wang2017emergent}. They showed that models which utilize explicit coreference information, for example via the attention sum mechanism (see Section \ref{sec:textcomp}), tend to perform better than those which do not. The suggested solution in that work was to add this information as extra features to the input of the model This approach forms our ``one-hot'' baseline. Here we take that idea further by proposing a modification to the structure of the reader itself for handling coreference, and show that doing so leads to further performance improvement, and interpretable output representations.
Our model is also related to the Recurrent Entity Networks architecture \citep{henaff2016tracking}. In fact, with perfect coreference information, each chain corresponds to an entity in the story, plus one chain for the sequential context. However, the MAGE-RNN model allows these chains to interact with each other, unlike recurrent entity networks (for example, the local context around the mention of an entity can inform the coreference representation of that entity and vice versa). It is also possible to incorporate other types of symbolic knowledge beyond coreference in MAGE-RNNs.


Recently, there has been interest in incorporating symbolic knowledge, such as that from a Knowledge Base or coreference information, within RNN-based language models \citep{yang2016reference,ahn2016neural}. However, rather than incorporating this knowledge within the structure of the RNN, these works instead use the output representation learned by the RNN to model latent variables which decide when to select the next token from the full vocabulary or a restricted vocabulary from the knowledge source. 
This approach is specific to the task of predicting the next token in a sequence. Our work is aimed at the more general problem of learning suitable representations of text.

\section{Methods}
\subsection{From Sequences to DAGs}
\label{sec:seqstodags}
Suppose that along with the input sequence $x_1, \ldots, x_T$, where $x_i \in \mathbb{R}^{d_{in}}$, we are also given information about which pairs of elements connect with each other. Further, suppose that these extra ``edges'' are typed---i.e., they belong to one of several different categories. Such extra information is common in Natural Language Processing (NLP). For example, one type of edge might connect multiple mentions of the same entity (coreference), while another type of edge might connect generic terms to their specific instances (hyponymy and hypernymy). Figure \ref{fig:dagexample} shows a simple example. Any piece of text can be augmented in this manner by running standard preprocessing tools such as coreference taggers and entity linkers.


Let $\mathcal{G} = (\mathcal{X}, \mathcal{E})$ denote the resulting directed graph which includes the sequential edges between consecutive elements, as well as the extra typed edges. The nodes $\mathcal{X} = \{x_i\}_{i=1}^T$ correspond to the elements of the original sequence, and edges $\mathcal{E}_0 = \{(x,x',e)\}$ are tuples consisting of the source, target and type of the link. The graph $\mathcal{G}$ results from augmenting the edges in $\mathcal{E}_0$ with inverse edges. More formally, for each edge $(x, x', e) \in \mathcal{E}_0$, we add an edge $(x', x, e')$ to the graph with $e'$ being the (artificial) inverse edge type of $e$. The resulting edge set with both original and inverse edges is denoted as $\mathcal{E}$. By definition, the graph $\mathcal{G}$ is a directed cyclic graph in general, but we can use the inherent order of the original sequence to decompose this into two subgraphs in the forward and backward directions respectively. The forward subgraph can be defined as $\mathcal{G}_f = (\mathcal{X}, \mathcal{E}_f)$, where $\mathcal{E}_f = \{(x_i,x_j,e_f)\in \mathcal{E}: i<j\}$. Here $i$ and $j$ are indices into the original sequence. The backward graph is defined analogously, with $\mathcal{E}_b = \{(x_i,x_j,e_b)\in \mathcal{E}: i>j\}$. By construction, $\mathcal{G}_f$ and $\mathcal{G}_b$ are DAGs, i.e., they do not contain cycles. We denote the set of all forward edge types by $E_f$ and all backward edge types by $E_b$.

For every DAG there exists a topological ordering of its nodes in a sequence such that all edges in the graph are directed from preceding nodes to succeeding nodes in the sequence. For $\mathcal{G}_f$ (respectively $\mathcal{G}_b$) defined above, one such ordering is immediately available -- the forward sequence order $(1,2,\ldots,T)$ (respectively the backward order $(T,T-1,\ldots,1)$). The existence of such an ordering makes DAGs particularly amenable to be modeled using RNNs, and below we discuss an architecture for doing so.

\subsection{MAGE-GRUs}


We present our framework as an adaptation of Gated Recurrent Units, called MAGE-GRU; however similar extensions can be derived for any recurrent neural network. 
MAGE-GRU uses separate networks for the forward and backward subgraphs respectively.
We present the updates only for the forward subgraph, since the backward subgraph is processed analogously.

\citet{miller2016key} and \citet{daniluk2017frustratingly} argue that overloaded use of state representations as both memory content and address makes training of the network difficult, and decompose these two functions by parameterizing them separately.
Similarly, we decompose the hidden states in the GRUs and maintain a separate hidden state vector $h^e_t \in \mathbb{R}^{d_e}$ for each edge type in $\mathcal{E}_f$.
The intuition behind this is that, for example, the representation flowing through the black edges in Figure \ref{fig:dagexample} need not be the same as that flowing through the red or green edges.

As $t$ varies from $1$ to $T$, the hidden states are updated in the topological order defined by the sequence and, importantly, the update for each edge state depends on the current state of \textit{all} incoming edges at $x_{t}$. Specifically, define
\begin{equation}
\mathcal{I}_f(x_t) = \{(t',e): (x_{t'}, x_t, e) \in \mathcal{E}_f\}
\end{equation}
as the set of incoming edges at node $x$, along with the index of their sources. 
Then the next state is given by
\begin{align}
r_t^e &= \sigma(W_r^e x_{t} + \sum_{(t',e') \in \mathcal{I}_f(x_t)} U_r^{e,e'} h^{e'}_{t'} + b^e_r) \nonumber \\
z_t^e &= \sigma(W_z^e x_{t} + \sum_{(t',e') \in \mathcal{I}_f(x_t)} U_z^{e,e'} h^{e'}_{t'} + b^e_z) \nonumber \\
\tilde{h}_t^e &= \text{tanh}(W_h^e x_{t} + r_t^e \odot \sum_{(t',e') \in \mathcal{I}_f(x_t)} U_h^{e,e'} h^{e'}_{t'} + b^e_h) \nonumber \\
h^e_t &= (1-z_t^e) \odot h^e_{t-1} + z_t^e \odot \tilde{h}_t^e
\label{eq:updates}
\end{align}
for each $e \in E_f$. $W_{r,z,h}^e$ and $U_{r,z,h}^{e,e'}$ are parameter matrices of size $d_e \times d_{in}$ and $d_e \times d_{e'}$ respectively, and $b_{r,z,h}^e$ are parameter vectors of size $d_e$. The output representation at time step $t$ is given by,
\begin{equation}
h_t = h_t^{e_1} \Vert h_t^{e_2} \Vert \ldots \Vert h_t^{e_{|E_f|}},
\label{eq:concat}
\end{equation}
where $h_t \in \mathbb{R}^{\sum_e d_e}$, and $\Vert$ denotes concatenation.
Output for the forward subgraph is given by $H_f = [h_1, \ldots, h_{T}]$. 
Similarly, we can obtain the output of the backward subgraph $H_b$, and concatenate with $H_f$ such that elements of the original sequence line up.
The collection of all previous output representations $M_t = [h_0; h_1; \ldots; h_{t-1}]$ can be viewed as the memory available to the 
recurrent model at time-step $t$.

\begin{figure}[t]
\includegraphics[width=\linewidth]{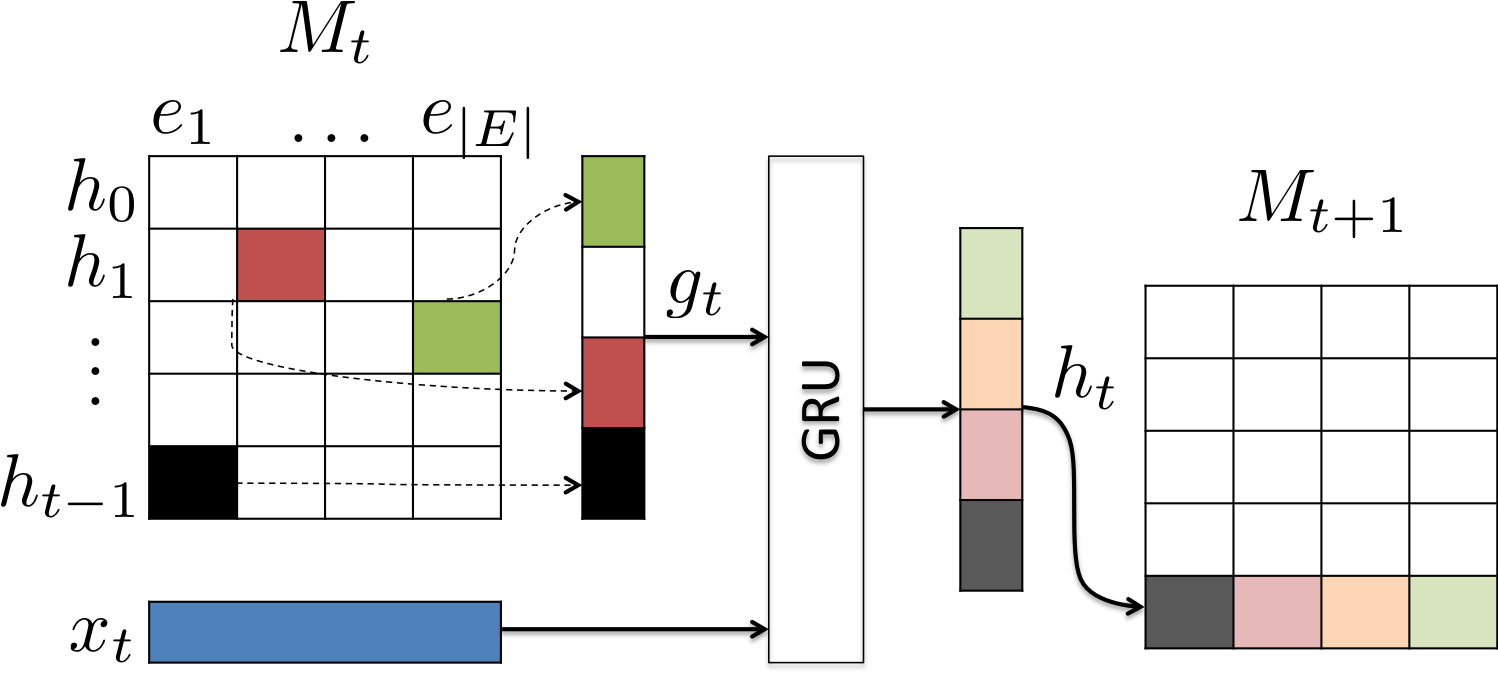}
\caption{Operation of the chain-decomposed MAGE-GRU. $M_t$ denotes the block of memories for each edge type and preceding time-steps.}
\label{fig:daggru}
\vspace{-0.2cm}
\end{figure}
In the case of coreference, or any relation where there is at most one incoming edge of a particular type at any node, the DAG can be decomposed into a collection of independent chains. Then the updates for each $e$ in (\ref{eq:updates}) can be trivially combined into one regular GRU update, as depicted in Figure \ref{fig:daggru}. To see this, define $g_t = g^{e_1}_t \Vert g^{e_2}_t \Vert \ldots \Vert g^{e_{|E|}}$, where $g^{e_i}_t = h_{t'}^{e_i}, \enskip \text{if} \enskip \exists (t',e_i) \in \mathcal{I}(x_t)$, else $g^{e_i}_t = 0$. In other words, $g_t^{e_i}$ holds the hidden state of the node connecting to $x_t$ via edge $e$, if there is such a node. Then, by stacking the matrices $W^e_*$, $U^{e,e'}_*$ and the vector $b^e_*$ for all $e, e'$, we obtain the following combined updates:
\begin{align}
r_t &= \sigma(W_r x_{s_t} + U_r g_t + b_r) \nonumber \\
z_t &= \sigma(W_z x_{s_t} + U_z g_t + b_z) \nonumber \\
\tilde{h}_t &= \text{tanh}(W_h x_{s_t} + r_t \odot U_h g_t + b_h) \nonumber \\
h_t &= (1-z_t) \odot h_{t-1} + z_t \odot \tilde{h}_t,
\label{eq:memupdates}
\end{align}
which are the usual GRU updates, but with the recurrent state replaced by $g_t$. Finally, we must slice the output vector $h_t$ back into its constituents $h^e_t$ for all $e \in E_f$ by extracting the appropriate dimensions. 

\subsection{Multiple Sequences}
\begin{figure}[t!]
    \centering
\includegraphics[width=0.8\linewidth]{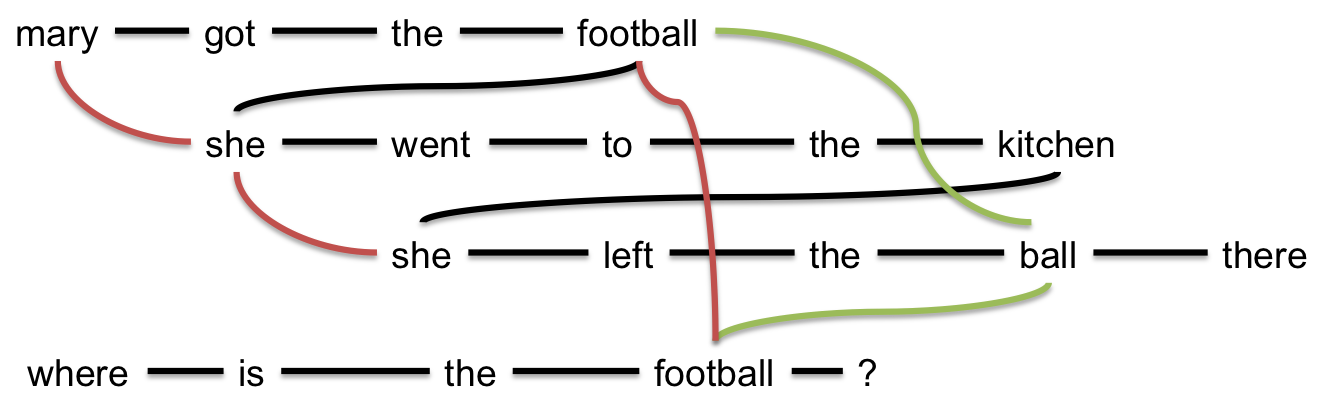}
\caption{Two sequences of text augmented with extra links. Red edges denote coreference relations, and green edges denote hypo/hyper-nymy. Black edges denote usual sequential links.}
\label{fig:multidagexample}
\vspace{-0.3cm}
\end{figure}
In certain applications, we have multiple sequences whose elements interact with each other via known relationships. Continuing with our motivating example, Figure \ref{fig:multidagexample} shows an example where the first sequence is a context passage and the second sequence is a question posed over the passage. The sequences are further augmented with coreference and hypernymy relations, resulting in an undirected cyclic graph. 

We would like to decompose this graph into a collection of DAGs and use the MAGE-GRU presented above to learn representations of the elements in the sequences. Also, we would like to preserve the order of the original sequences in the decomposed DAGs. Suppose we have $S$ sequences $X_1, \ldots, X_S$. One way to do this is as follows: for each permutation of the collection of sequences $(X_{k_1}, X_{k_2}, \ldots, X_{k_S})$, treat it as a single long sequence and decompose into forward and backward subgraphs as described in Section \ref{sec:seqstodags}. However, this results in $2S!$ DAGs, which is expensive except for very small $S$. Instead, we propose here taking one random permutation of the sequences and decomposing it into the forward and backward subgraphs. In this manner, each edge in the graph is still traversed twice (once in both directions), without incurring any additional cost compared to processing the sequences individually. Moreover, multi-layer extensions of the MAGE-GRU can allow information to flow through arbitrary paths through the graph. 

\section{Experiments} \label{sec:exp}
\subsection{Text Comprehension with Coreference}
\label{sec:textcomp}
Text comprehension tasks involve tuples of the form $(d,q,a)$, where $d$ is a context document and $q$ is a question whose answer $a$ can be inferred from reading $d$. In the \textit{extractive} setting the answer is a span of text in the context document. Often a set of candidates $\mathcal{C}$ is available, in which case the task becomes a \textit{classification} task. Several large-scale benchmarks \citep{rajpurkar2016squad,onishi2016did} and deep learning models \citep{munkhdalai2016reasoning} have been proposed for this task recently.

Comprehending a document requires complex reasoning processes and prior knowledge on the part of the reader. One of these processes is \textit{coreference resolution}, where the reader must identify expressions in text referring to the same entity (see red edges in Figures \ref{fig:dagexample}, \ref{fig:multidagexample}). \citet{chu2016broad} show that existing state-of-the art models have poor performance in cases where some form of coreference resolution is required. A separate line of work \citep{clark2015entity}, however, has led to the development of sophisticated coreference annotators\footnote{\url{http://stanfordnlp.github.io/CoreNLP/coref.html}}, and here we use these annotations over document and question pairs as an explicit memory signal for an RNN reader. The intuition is that when processing referring expressions, the model should also receive information about previous mentions of the referent entity. 

We study the effect of adding an explicit coreference signal to RNN based models for three well studied text comprehension benchmarks. 
Following previous work \citep{hermann2015teaching,kadlec2016text} our basic architecture consists of bidirectional GRUs to encode the document and query into a matrix $H^d = [h^d_1,\ldots,h^d_{|d|}]$ and vector $h^q$ respectively. 
Next the query vector is used to compute an attention distribution over the document,
\begin{equation}
    \alpha = \text{softmax}\left( h^q \right)^T H^d
\end{equation}
For extractive tasks, we use this attention distribution directly to predict the answer, using the \textit{attention-sum} mechanism suggested by \citet{kadlec2016text}. Hence, the probability of selecting token $w$ as the answer is given by $\sum_{i \in I(w,d)} \alpha_i$, where $I(w,d)$ is the set of positions $w$ occurs in $d$. For classification tasks, we select the answer among the candidates $\mathcal{C}$ as follows:
\begin{align}
    h^d &= \alpha^T H^d \nonumber \\
    p_{\mathcal{C}} &= \text{softmax}((h^{d})^T W_{\mathcal{C}}) \nonumber \\ 
    \hat{a} &= \text{argmax}_{\mathcal{C}} p_{\mathcal{C}} \label{eq:attend}
\end{align}
where $W_{\mathcal{C}}$ is a lookup table of output embeddings, one for each candidate.

To test our contribution, we replace the pair of bi-GRUs with the single MAGE-GRU model described for multiple sequences for
computing the document and query representations, and compare the final performance.
As a baseline we also compare to the setting where coreference information is added as extra features at the input of the GRU. 
Let $M$ be the number of coreference chains for the $(d,q)$ pair: we append a one-hot vector $o_t \in \{0,1\}^M$ to the input of the 
GRU $x_t$ indicating which coreference chain, if any, that token is a part of. Such additional features were shown to be useful by \citet{wang2017emergent}.
Henceforth, we refer to this baseline as ``one-hot''.

\citet{dhingra2016gated} presented a multi-layer architecture called \textit{GA Reader} for text comprehension which achieves state of the art performance
over several datasets. To further test whether our model can improve over this competitive baseline, we replace the bi-GRU at every layer of the 
GA Reader with our proposed MAGE-GRU.

\subsection{Performance Comparison}
\begin{table*}[htbp!]
\small
\centering
\renewcommand{\arraystretch}{1.0}
\begin{tabular}{@{}lRRRRRRRRRRRRRRRRRRRRRR@{}}
    \toprule
    \multirow{2}{*}{\textbf{Model}}      & \multicolumn{20}{c}{\textbf{Task}}                                                                                                                                                                                                                                                         & \multicolumn{2}{c}{\textbf{Overall}} \\ \cmidrule(l){2-23} 
                                         & \textbf{1} & \textbf{2} & \textbf{3} & \textbf{4} & \textbf{5} & \textbf{6} & \textbf{7} & \textbf{8} & \textbf{9} & \textbf{10} & \textbf{11} & \textbf{12} & \textbf{13} & \textbf{14} & \textbf{15} & \textbf{16} & \textbf{17} & \textbf{18} & \textbf{19} & \textbf{20}               & \textbf{Avg}   & \textbf{\#failed}   \\ \midrule
N2N             & 0.1        & 18.8       & 31.7       & 17.5       & 12.9       & 2.0        & 10.1       & 6.1        & 1.5        & 2.6         & 3.3         & 0.0         & 0.5         & 2.0         & 1.8         & 51.0        & 42.6        & 9.2         & 90.6        & 0.2  & 15.2           & 10                  \\
QRN             & 0.0        & 0.7        & 5.7        & 0.0        & 1.1        & 0.9        & 9.6        & 5.6        & 0.0        & 0.0         & 0.0         & 0.0         & 0.0         & 0.8         & 0.0         & 53.0        & 34.4        & 7.9         & 78.7        & 0.2  & 9.9            & 7                   \\ \midrule
bi-GRU          & 0.0        & 59.7       & 51.1       & 0.5        & 2.4        & 21.5       & 18.7       & 23.3       & 23.4       & 21.8        & 0.6         & 0.4         & 0.8         & 20.5        & 50.4        & 51.5        & 35.7        & 8.8         & 81.5        & 1.8  & 23.7           & 13                  \\
+one-hot        & 0.4        & 65.7       & 51.8       & 0.3        & 1.3        & 33.9       & 22.4       & 24.3       & 23.7       & 17.8        & 0.5         & 0.9         & 0.7         & 21.6        & 46.4        & 51.0        & 42.7        & 9.9         & 80.5        & 2.3  & 24.9           & 13                  \\
+MAGE (16)      & 0.0        & 9.7        & 1.5        & 0.0        & 1.5        & 3.0        & 12.0       & 11.9       & 1.2        & 0.2         & 0.0         & 0.1         & 1.9         & 9.7         & 27.3        & 45.0        & 33.8        & 8.3         & 37.7        & 0.2  & 10.3           & 9                   \\ \midrule
GA              & 0.0        & 62.4       & 47.3       & 0.8        & 1.4        & 18.3       & 16.9       & 23.7       & 20.5       & 20.2        & 0.0         & 0.0         & 0.6         & 23.9        & 51.3        & 50.8        & 42.1        & 7.1         & 91.3        & 3.6 & 24.1           & 13                  \\
+ one-hot       & 0.0        & 66.4       & 49.3       & 0.5        & 1.1        & 21.0       & 16.0       & 15.6       & 25.8       & 33.6        & 0.2         & 0.0         & 0.9         & 22.6        & 52.0        & 50.7        & 39.6        & 10.0        & 90.5        & 3.9  & 25.0           & 13                  \\
+ DAG-RNN       & 0.0        & 22.9       & 46.3       & 0.4        & 2.5        & 9.1        & 7.5        & 46.4       & 0.8        & 0.6         & 2.0         & 3.4         & 5.9         & 19.5        & 63.3        & 51.4        & 46.0        & 7.4         & 92.1        & 7.8  & 21.8           & 13                  \\
+ MAGE (shared) & 0.0        & 1.2        & 2.6        & 0.0        & 0.9        & 1.7        & 6.3        & 9.6        & 0.1        & 0.7         & 0.0         & 0.1         & 0.0         & 19.7        & 34.7        & 3.0         & 43.3        & 9.9         & 62.8        & 0.1  & 9.8            & 7                   \\
+ MAGE (16)     & 0.0        & 1.0        & 1.5        & 0.0        & 0.6        & 0.5        & 4.8        & 2.5        & 0.2        & 0.4         & 0.1         & 0.0         & 0.6         & 13.8        & 18.4        & 2.4         & 43.2        & 5.5         & 76.0        & 2.4  & \textbf{8.7}   & \textbf{5}          \\ \bottomrule
\end{tabular}
\caption{Error rates on the $20$ bAbi tasks. N2N refers to End-to-end Memory Networks \citep{sukhbaatar2015end}. QRN refers to Query Reduction Networks \citep{seo2016query}. For MAGE models a number in the parenthesis indicates the dimensions for coreference states in the split output. ``shared'' in the parenthesis indicates that outputs were shared for all edge types. DAG-RNN refers to the model proposed by \citet{shuai2016dag} incorporated with GA. ``one-hot'' refers to the baseline where coreference features are appended to the input. A task is considered failed if the error rate is above $5\%$.}
\label{tab:babi}
\vspace{-0.3cm}
\end{table*}

\begin{table}[t]
    \small
    \centering
    \begin{tabular}{lll}
        \toprule
        \textbf{Model} & \textbf{Average} & \textbf{Task 3} \\ \midrule
        QRN & 17.1 & 72.8 \\
        MAGE (shared) & \textbf{11.8} & 1.2 \\
        MAGE (16) & 14.1 & \textbf{0.3} \\ \bottomrule
    \end{tabular}
    \caption{Performance comparison on bAbi-Mix. We report average error rate over 20 tasks, as well as the error rate on Task 3.}
    \label{tab:mixe}
    \vspace{-0.5cm}
\end{table}

\paragraph{Story Based QA:} Our first benchmark is the bAbi dataset from \citet{weston2015towards}, a set of $20$ toy tasks aimed at measuring the ability of agents to reason about natural language. In each task the agent is presented with a simple story about entities in operating in an environment, followed by a question about the final state of that environment. Different tasks measure different reasoning abilities on the part of the agent, including chaining facts, counting, deduction, induction and others. An example\footnote{the actual dataset only contains named mentions and not pronouns. These were introduced in the figure for exposition.} is shown in Figure \ref{fig:multidagexample}. The official release of the dataset\footnote{\url{https://research.fb.com/downloads/babi/}} comes with two versions, we focus on the more difficult 1K split in this work.
Following \citet{seo2016query}, we ran $10$ random initializations of the model and report the test set performance for the model with the best validation set performance. 

The natural language in the stories was generated using templates, which are easy to learner.
Rather, the difficulty of these tasks lies in tracking the state of multiple entities across long distances in the input sequence, 
which makes them particularly suitable for testing our proposed model with explicit memory. 
Specifically, we connect consecutive mentions of the same entity with coreference links, 
which are easily extracted due to the synthetic nature of the language. 
Table~\ref{tab:babi} shows a comparison of previous works with our proposed models and several baselines.
Our model achieves new state-of-the-art results, outperforming strong baselines such as QRNs. Moreover, we observe that the proposed MAGE architecture can substantially improve the performance for both bi-GRUs and GAs. Adding the same information as one-hot features fails to improve the performance, which indicates that the inductive bias we employ on MAGE is useful.
The DAG-RNN baseline from \citet{shuai2016dag} and the shared version of MAGE (where edge representations are tied) also perform worse, showing that our proposed 
architecture is superior.

Our motivation in incorporating the DAG structure in text comprehension models is to help the reader model long term dependencies in the input sequences. 
To test how well it is able to do that, we constructed a new version of the bAbi tasks, called \textit{bAbi-Mix}, where each story is a mixed version
of two independent stories in two different worlds. This was done by first replacing all entity mentions in one of the stories with alternates, so 
``Daniel'' becomes ``David'', ``milk'' becomes ``juice'' and so on. Then we mixed the sentences in the two stories, in random order, and asked a question about
one of the stories. As a result, the stories become longer, and relevant information about entities has to be tracked over longer distances. Answering the questions
is still trivial for human readers, but becomes even more challenging for RNN models. Table~\ref{tab:mixe} shows the performance of QRNs and MAGEs. Both variants of MAGE substantially outperform QRNs, which are the current state-of-the-art models on the bAbi dataset. As a case study, we also report the performances on Task 3 which requires reasoning over three supporting facts. We observe that MAGE increases the ability to model multiple facts and reduces the error rate from 72.8\% to 0.3\%.

\begin{figure*}[t]
\centering
\subfigure{\label{fig:babi7}\includegraphics[width=85mm,trim={1.5cm 0.5cm 2.5cm 0cm},clip]{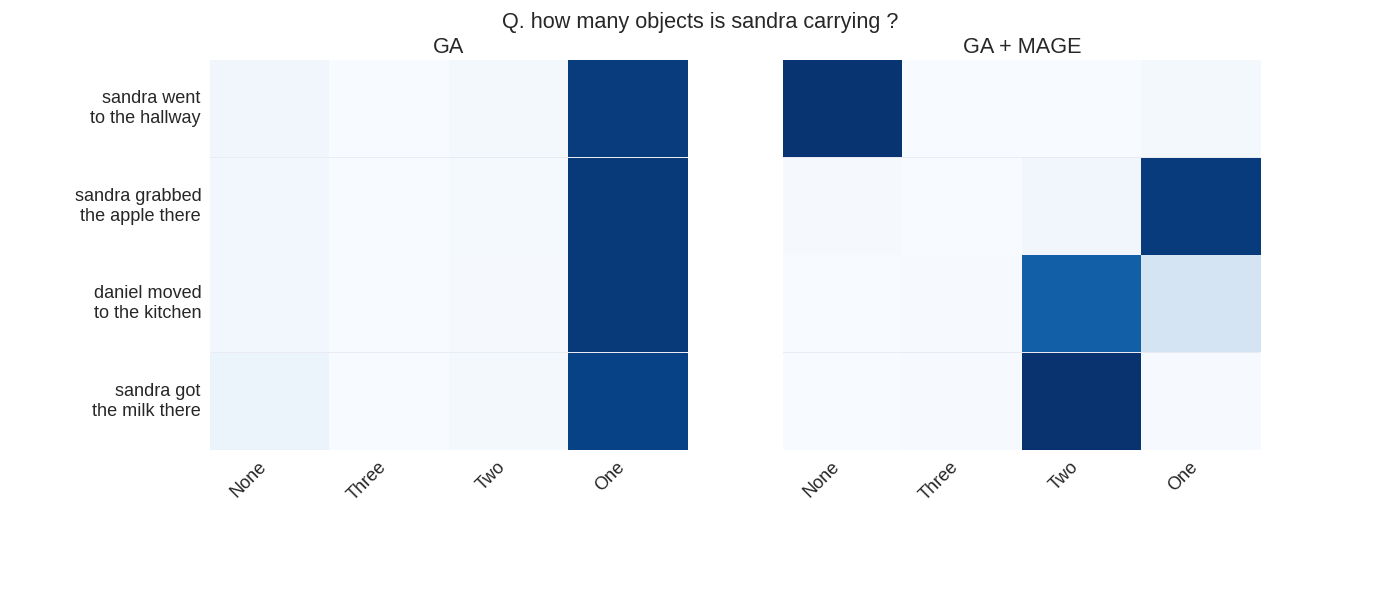}}~
\subfigure{\label{fig:babi8}\includegraphics[width=85mm,trim={1.5cm 0.5cm 2.5cm 0cm},clip]{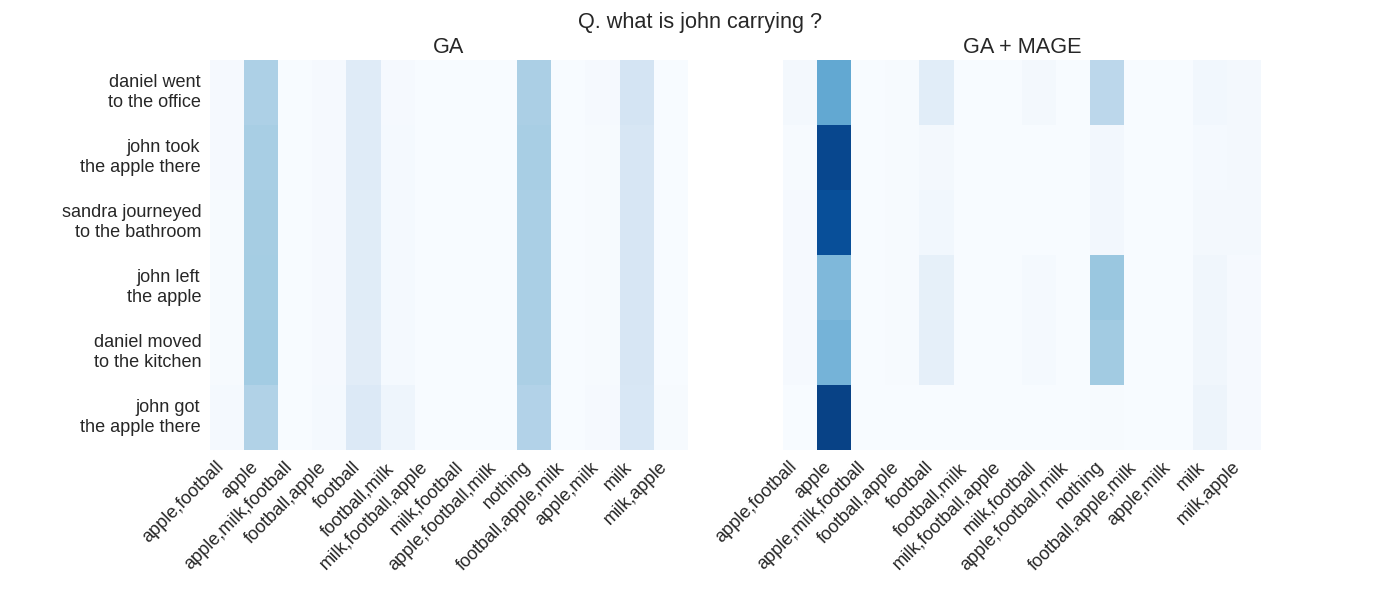}}
\vspace{-8mm}
\caption{Heat map of the distribution over candidate answers computed from the document representation after each sentence (shown on the y-axis) in the story. This is done by replacing the attention weighted represention $h^d$ in Eq. \ref{eq:attend} with the one at the sentence boundary. Darker values indicate higher scores. The left example is from Babi Task 7  and the right from Task 8. We compare both GA and GA+MAGE models.}
\label{fig:visual}
\vspace{-0.3cm}
\end{figure*}
To further gain insight into the representations being learned by the models, we visualize the scores assigned to candidate answers at intermediate points in the story 
by GA and GA+MAGE in Figure~\ref{fig:visual}. We picked the document representation at sentence boundaries, and computed its inner product with the output lookup
table $W_{\mathcal{C}}$ followed by the softmax nonlinearity. The resulting distribution for two such stories is plotted in each row of Figure~\ref{fig:visual}. In
these examples, and many more that we analyzed, we observed that the output distribution remains more or less constant across the story. Hence, the model tries to learn a 
global representation which selects the correct answer, sacrificing the fine-grained information of what is happening in the story at intermediate points.
In contrast, the output distribution for GA+MAGE accurately reflects the state of the story at the point that it is computed. Indeed, the biggest gains for GA+MAGE over GA 
are in tasks which require tracking the state of entities across the story, something that the learned representation seems to encode well.
This could be potentially useful in a situation where the agent is required to answer multiple questions about a story.

\begin{table}[t]
    \centering
    \small
    \begin{tabular}{@{}lcccc@{}}
            \toprule
            \multirow{2}{*}{Lambada}  & \multicolumn{2}{c}{context}     & \multicolumn{2}{c}{all}         \\ \cmidrule(l){2-5} 
            & val            & test           & val            & test           \\ \midrule
            n-gram LM$\dagger$        & --             & --             & --             & 0.118          \\
            GA + features $\dagger$ & --             & --             & --             & 0.490          \\ \midrule
            bi-GRU                     & 0.550          & 0.556          & 0.445          & 0.451          \\
            + one-hot                 & 0.558          & 0.555          & 0.451          & 0.450          \\
            + MAGE (48)                & 0.574          & 0.573          & 0.464          & 0.465          \\ \midrule
            GA                        & 0.611          & 0.616          & 0.494          & 0.500          \\
            + one-hot                 & 0.633          & 0.634          & 0.512          & 0.515          \\
            + MAGE (48)                & 0.632          & \textbf{0.636} & 0.511          & \textbf{0.516} \\
            + MAGE (64)                & \textbf{0.644} & 0.630          & \textbf{0.521} & 0.511          \\ \midrule
            Upper Bound               & 1              & 1              & 0.808          & 0.811          \\
            Human$\dagger$         & --             & --             & --             & 0.86           \\ \bottomrule
        \end{tabular}
        \caption{Validation and Test set accuracies on LAMBADA dataset for splits where answer is in the passage (\textit{context}) and full set (\textit{all}). ``MAGE'' refers to our proposed model, where the number within parentheses denotes the number of hidden dimensions for coreference edges. ``one-hot'' refers to a model where coreference ids are appended to the input word embeddings. Results marked with $\dagger$ are cf \citep{chu2016broad}. Upper bound is the best performance a system extracting answers from the passage can achieve. Human performance was estimated from $100$ randomly sampled instances.}
        \label{tab:lambada}
        \vspace{-0.5cm}
\end{table}

\paragraph{Broad Context Language Modeling:} For our second benchmark we pick the LAMBADA dataset from \citet{paperno2016lambada}, where the task is to predict the last word in a given passage. The passages are filtered such that human subjects were able to predict the last word when given a broad context of 4-5 sentences, but not when only given the immediately preceding sentence. This filtering process makes the task a challenging one -- any model must understand the broader discourse to predict the correct word. The passages here are selected from fiction books, and hence language comprehension is challenging, unlike the bAbi tasks. Standard language models only achieve about $7.3\%$ accuracy on this dataset, but \citet{chu2016broad} improved this to $49\%$ by reformulating the task as a reading comprehension one, and training on a large corpus of automatically extracted passages. They treat the last sentence in the passage as the query and the remaining passage as context document, and extract the answer from this context instead. This limits the resulting model to only predict words when they are in the context, which is the case for $81\%$ of passages in the test set.

\citet{chu2016broad} also performed manual analysis on a subset of the test set and found that approximately $20\%$ of the passages required coreference resolution to find the correct answer. With this motivation, we extracted coreference chains for each passage in the dataset using the Stanford CoreNLP tools\footnote{\url{http://stanfordnlp.github.io/CoreNLP/coref.html}}, and compare the performance of baseline models with our proposed MAGE-GRU, listed in Table~\ref{tab:lambada}. We keep the total hidden state size, and hence number of parameters, fixed for all models at $256$, but for MAGE models part of this is allocated to sequential edges and part of it is allocated to coreference edges. We focus on the split where the answer is in context, since the architecture only attempts to solve these passages. Our implementation of GA gave higher performance than that reported by \cite{chu2016broad}, without the use of linguistic features. We believe the difference is because we use a newer release of the code by the authors of GA \cite{dhingra2016gated}.

On the simple bi-GRU architecture we see an improvement of $1.7\%$ by incorporating coreference edges in the graph, whereas the one-hot baseline does not lead to any improvement. Hence, simply providing the model information about which tokens in text refer to which entity is not enough; modifying the structure of the network to allow it to propagate these memories is important. On the multi-layer GA architecture, the coreference edges again lead to an improvement of $2\%$, setting a new state-of-the-art on this dataset. In this case we see that the one-hot baseline also performs comparably. This suggests that for short documents (LAMBADA passages have an average size of $75$ tokens), multi-layer architectures are able to propagate reference information when given an explicit input signal.

\begin{table}[t]
    \small
    \centering
    \begin{tabular}{@{}lccc@{}}
        \toprule
        \textbf{label}           & \textbf{\#} & \textbf{GA} & \textbf{\begin{tabular}[c]{@{}c@{}}GA +\\ MAGE\end{tabular}} \\ \midrule
        single name cue          & 9           & 0.67        & 1.00                                                        \\
        simple speaker tracking  & 19          & 0.74        & 0.79                                                        \\
        discourse inference rule & 16          & 0.63        & 0.75                                                        \\
        coreference              & 21          & 0.48        & 0.62                                                        \\
        basic reference          & 18          & 0.50        & 0.61                                                        \\
        semantic trigger         & 20          & 0.40        & 0.55                                                        \\
        external knowledge       & 24          & 0.33        & 0.50                                                        \\ \midrule
        all                      & 100         & 0.51        & 0.61                                                        \\ \bottomrule
    \end{tabular}
    \caption{Performance of the GA and GA+MAGE architectures for each category of manually annotated labels. Please see \citet{chu2016broad} for a detailed description of the labels.}
    \label{tab:manual}
    \vspace{-0.5cm}
\end{table}
Table~\ref{tab:manual} shows a comparison of the baseline GA architecture with the coreference augmented GA+MAGE model on the $100$ manually labeled validation instances available from \citet{chu2016broad}. The small sample size for each category makes it hard to draw strong conclusions from these results. Nevertheless, we note that GA+MAGE consistently outperforms GA in each category, with the biggest improvements coming for the \textit{single name cue}, \textit{semantic trigger}, \textit{coreference} and \textit{external knowledge} labels.

\begin{figure}[t]
\centering
\includegraphics[width=\linewidth]{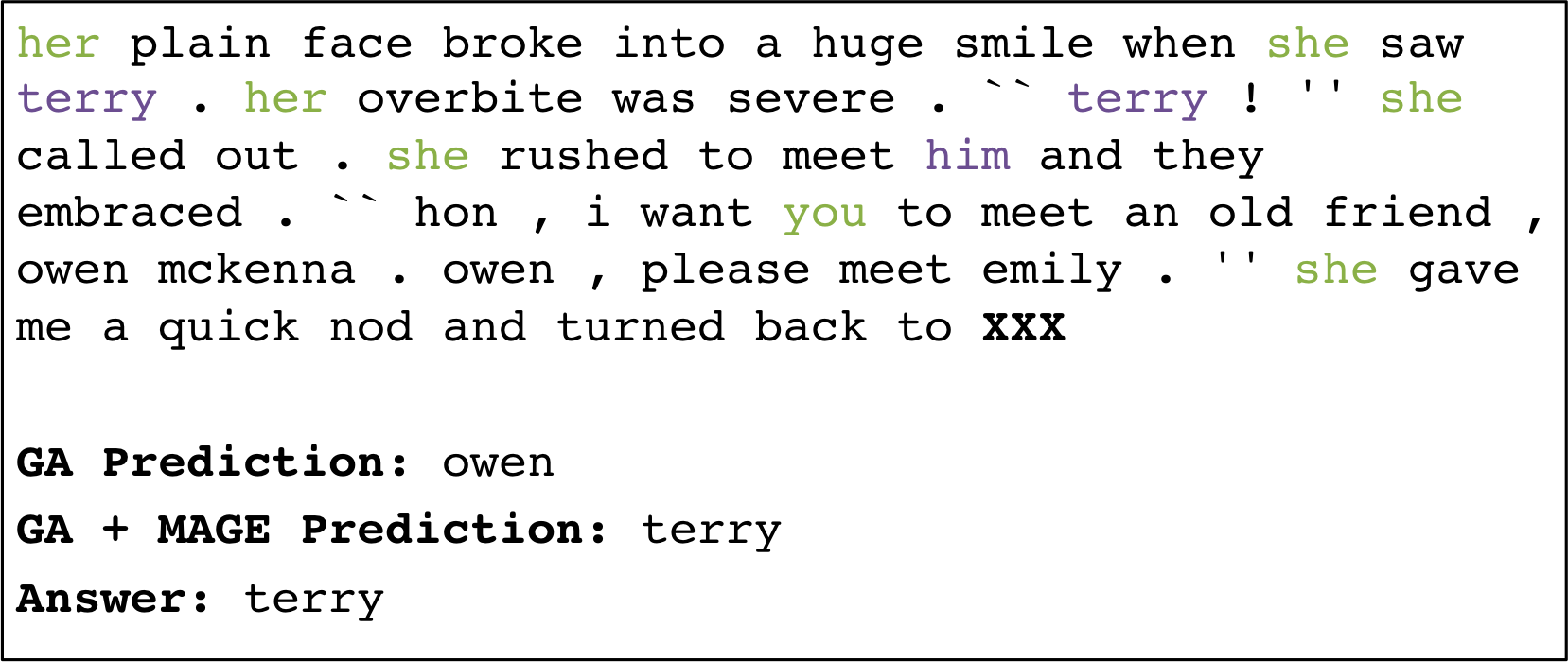}\\
\includegraphics[width=\linewidth]{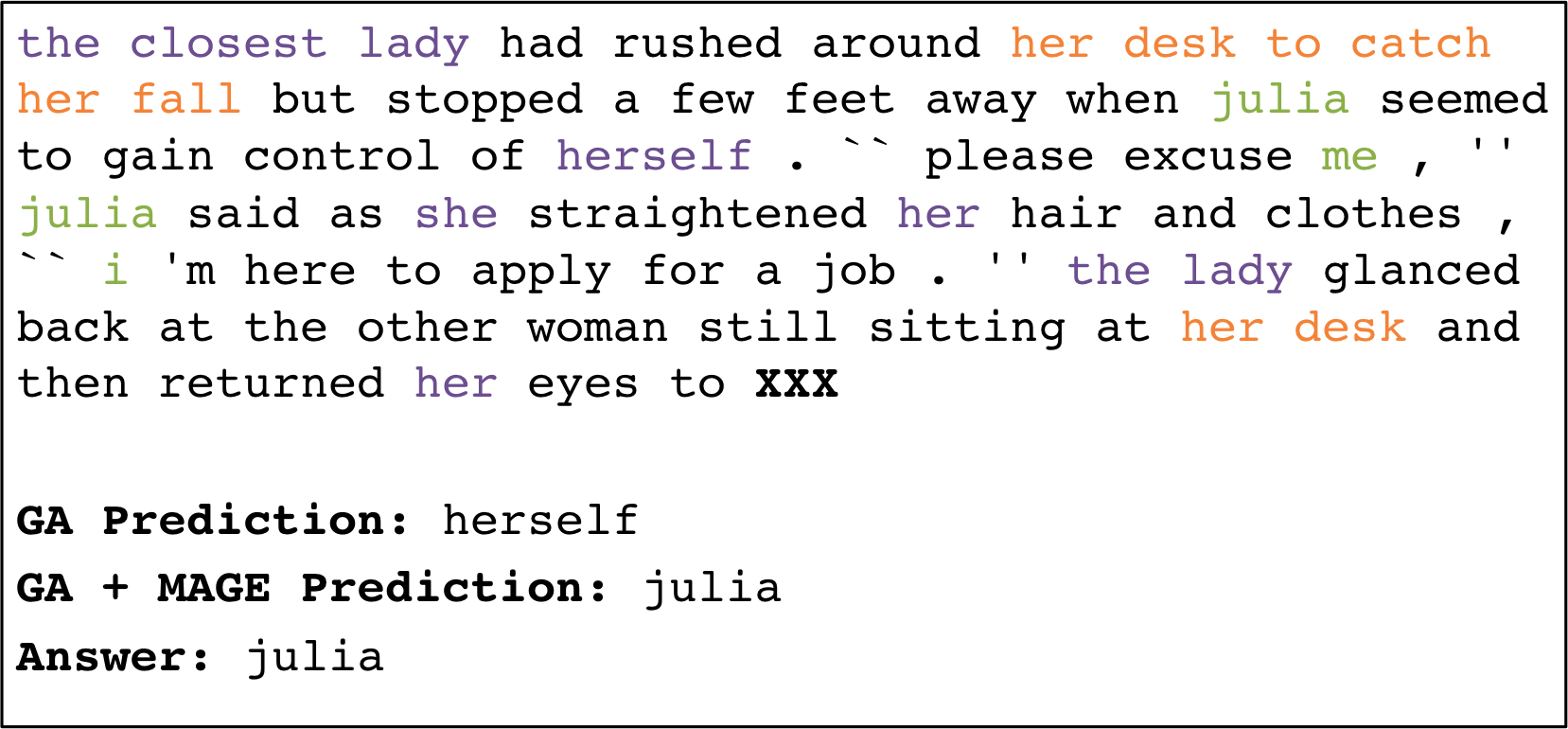}
\vspace{-4mm}
\caption{Two example passages from the LAMBADA dataset with predictions from the GA and GA+MAGE models. Tokens of the same color represent coreferent mentions of the same entity, as extracted by the preprocessing tool.}
\label{fig:lambada-visual}
\vspace{-0.2cm}
\end{figure}
Figure~\ref{fig:lambada-visual} shows passages from the LAMBADA dataset along with the annotated coreferences and the predictions from GA and GA+MAGE models. In both cases GA predicts the wrong entities as the answer. Instead MAGE, which is able to track entity states with the provided coreference signals, answers both passages correctly.

\begin{table}[t!]
\small
\centering
\begin{tabular}{@{}lcc@{}}
    \toprule
    \textbf{CNN} & \textbf{val}   & \textbf{test}  \\ \midrule
BiDAF$\dagger$        & 0.763          & 0.769          \\ \midrule
BiGRU        & 0.694          & 0.704          \\
+ one-hot    & 0.692          & 0.701          \\
+ MAGE (48)  & 0.722          & 0.729          \\ \midrule
GA$\ddagger$           & 0.779          & 0.779          \\
+ one-hot    & 0.780          & 0.781 \\
+ MAGE (16)  & 0.792          & 0.781          \\
+ MAGE (32)  & \textbf{0.792} & \textbf{0.786}          \\ \bottomrule
\end{tabular}
\caption{Validation and Test set accuracies on CNN dataset. ``MAGE'' refers to our proposed model, where the number within parentheses denotes the size of coreference states. ``one-hot'' refers to a model where coreference ids are appended to the input word embeddings. ``BiDAF'' refers to the Bidirectional Attention Flow model. Results marked with $\dagger$ are cf \citep{seo2016bidirectional} and with $\ddagger$ from \citep{dhingra2016gated}.}
\label{tab:cnn}
\vspace{-0.3cm}
\end{table}
\paragraph{Cloze-style QA:} Lastly, we test our models on the CNN dataset from \citet{hermann2015teaching}, which consists of pairs of news articles and a cloze-style question over the contents of the article\footnote{The queries are constructed by replacing an entity in the summary of the article with a placeholder, and the task is to find the entity}. Since its release, the dataset has recieved much attention and several deep learning architectures have been introduced with impressive results. An interesting property of the dataset is that in each article named entities and expressions referring to them have been anonymized by replacing with placeholders (such as \texttt{@entity24}) to make the task purely a comprehension one. For our purposes, without the use of any external tools, we augment the article with extra edges connecting mentions of the same entity, and compare the performance with and without these links. Table~\ref{tab:cnn} shows the performance comparison. Augmenting the bi-GRU model with MAGE leads to an improvement of $2.5\%$ on the test set. The previous best results for this dataset were achieved by the GA Reader, and we see that adding MAGE to it leads to a further improvement of $0.7\%$, setting a new state of the art. This is an impressive improvement, given that previous works have reported that we are already close to the upper bound possible on this dataset \citep{chen2016thorough}. Note that we are not adding any information beyond what is already available in the dataset, since entity mentions are anonymized. 


\section{Conclusions}
We have presented a framework for incorporating symbolic knowledge such as linguistic relations between tokens
in text into recurrent neural networks. We interpret these relations as an explicit memory signal and augment
the chain structured RNN with edges connecting the arguments of the relations.
Our model, MAGE-RNN, parameterizes each edge type separately, and also maintains a separate hidden state representation
for distinct edges at every node.
It can be interpreted as a memory-augmented RNN where the memory access
is dictated by the graph structure.
We apply the MAGE-RNN framework to model coreference for text comprehension tasks by 
preprocessing to extract coreference relations and replacing recurrent units in comprehension models with MAGE-RNN.
We observe consistent improvements across three widely studied benchmarks, for both simple and sophisticated
architectures. Our best results set a new state of the art on all three tasks.

The ultimate goal in machine learning is, of course, to be able to learn purely 
from data, without relying on external tools. However, in practice this is only feasible when the training dataset
size is large, which is often not the case in NLP applications. We hypothesize, however, that the biggest benefit of explicit
linguistic knowledge will be in cases where the data is scarce, for example as we observe in the bAbi tasks. 
Moreover, since coreference and entity-linking tools are widely available, models which exploit them are valuable.
Hence, an interesting future direction of this work is
to incorporate an attention mechanism over the edge types into MAGE-RNN and study its distribution over various sources
as the dataset size varies.
Coreference is one important type of linguistic knowledge that machine comprehension models can benefit from.
Our encouraging results motivate us to explore other potentially useful sources of knowledge, which
may include -- dependency parses, semantic role labels, semantic frames,
ontologies such as Wordnet \citep{miller1990introduction}, and databases 
such as Freebase \citep{bollacker2008freebase}. MAGE-RNN is a general framework capable of integrating
all these sources into a single neural model, and we plan to investigate this research in future work.

\section*{Acknowledgments}
This work was funded by NSF under CCF1414030 and Google Research.

\bibliography{example_paper}

\begin{thebibliography}{37}
\providecommand{\natexlab}[1]{#1}
\providecommand{\url}[1]{\texttt{#1}}
\expandafter\ifx\csname urlstyle\endcsname\relax
  \providecommand{\doi}[1]{doi: #1}\else
  \providecommand{\doi}{doi: \begingroup \urlstyle{rm}\Url}\fi

\bibitem[Ahn et~al.(2016)Ahn, Choi, P{\"a}rnamaa, and Bengio]{ahn2016neural}
Ahn, Sungjin, Choi, Heeyoul, P{\"a}rnamaa, Tanel, and Bengio, Yoshua.
\newblock A neural knowledge language model.
\newblock \emph{arXiv preprint arXiv:1608.00318}, 2016.

\bibitem[Bahdanau et~al.(2014)Bahdanau, Cho, and Bengio]{bahdanau2014neural}
Bahdanau, Dzmitry, Cho, Kyunghyun, and Bengio, Yoshua.
\newblock Neural machine translation by jointly learning to align and
  translate.
\newblock \emph{arXiv preprint arXiv:1409.0473}, 2014.

\bibitem[Bengio et~al.(1994)Bengio, Simard, and Frasconi]{bengio1994learning}
Bengio, Yoshua, Simard, Patrice, and Frasconi, Paolo.
\newblock Learning long-term dependencies with gradient descent is difficult.
\newblock \emph{IEEE transactions on neural networks}, 5\penalty0 (2):\penalty0
  157--166, 1994.

\bibitem[Bollacker et~al.(2008)Bollacker, Evans, Paritosh, Sturge, and
  Taylor]{bollacker2008freebase}
Bollacker, Kurt, Evans, Colin, Paritosh, Praveen, Sturge, Tim, and Taylor,
  Jamie.
\newblock Freebase: a collaboratively created graph database for structuring
  human knowledge.
\newblock In \emph{Proceedings of the 2008 ACM SIGMOD international conference
  on Management of data}, pp.\  1247--1250. AcM, 2008.

\bibitem[Chen et~al.(2016)Chen, Bolton, and Manning]{chen2016thorough}
Chen, Danqi, Bolton, Jason, and Manning, Christopher~D.
\newblock A thorough examination of the cnn/daily mail reading comprehension
  task.
\newblock \emph{ACL}, 2016.

\bibitem[Cho et~al.(2014{\natexlab{a}})Cho, Van~Merri{\"e}nboer, Bahdanau, and
  Bengio]{cho2014properties}
Cho, Kyunghyun, Van~Merri{\"e}nboer, Bart, Bahdanau, Dzmitry, and Bengio,
  Yoshua.
\newblock On the properties of neural machine translation: Encoder-decoder
  approaches.
\newblock \emph{arXiv preprint arXiv:1409.1259}, 2014{\natexlab{a}}.

\bibitem[Cho et~al.(2014{\natexlab{b}})Cho, Van~Merri{\"e}nboer, Gulcehre,
  Bahdanau, Bougares, Schwenk, and Bengio]{cho2014learning}
Cho, Kyunghyun, Van~Merri{\"e}nboer, Bart, Gulcehre, Caglar, Bahdanau, Dzmitry,
  Bougares, Fethi, Schwenk, Holger, and Bengio, Yoshua.
\newblock Learning phrase representations using rnn encoder-decoder for
  statistical machine translation.
\newblock \emph{arXiv preprint arXiv:1406.1078}, 2014{\natexlab{b}}.

\bibitem[Chu et~al.(2016)Chu, Wang, Gimpel, and McAllester]{chu2016broad}
Chu, Zewei, Wang, Hai, Gimpel, Kevin, and McAllester, David.
\newblock Broad context language modeling as reading comprehension.
\newblock \emph{arXiv preprint arXiv:1610.08431}, 2016.

\bibitem[Clark \& Manning(2015)Clark and Manning]{clark2015entity}
Clark, Kevin and Manning, Christopher~D.
\newblock Entity-centric coreference resolution with model stacking.
\newblock In \emph{ACL (1)}, pp.\  1405--1415, 2015.

\bibitem[Daniluk et~al.(2017)Daniluk, Rocktäschel, Welbl, and
  Riedel]{daniluk2017frustratingly}
Daniluk, Michał, Rocktäschel, Tim, Welbl, Johannes, and Riedel, Sebastian.
\newblock Frustratingly short attention spans in neural language modeling.
\newblock \emph{arXiv preprint arXiv:1702.04521}, 2017.

\bibitem[Dhingra et~al.(2016)Dhingra, Liu, Cohen, and
  Salakhutdinov]{dhingra2016gated}
Dhingra, Bhuwan, Liu, Hanxiao, Cohen, William~W, and Salakhutdinov, Ruslan.
\newblock Gated-attention readers for text comprehension.
\newblock \emph{arXiv preprint arXiv:1606.01549}, 2016.

\bibitem[Graves et~al.(2014)Graves, Wayne, and Danihelka]{graves2014neural}
Graves, Alex, Wayne, Greg, and Danihelka, Ivo.
\newblock Neural turing machines.
\newblock \emph{arXiv preprint arXiv:1410.5401}, 2014.

\bibitem[Henaff et~al.(2016)Henaff, Weston, Szlam, Bordes, and
  LeCun]{henaff2016tracking}
Henaff, Mikael, Weston, Jason, Szlam, Arthur, Bordes, Antoine, and LeCun, Yann.
\newblock Tracking the world state with recurrent entity networks.
\newblock \emph{arXiv preprint arXiv:1612.03969}, 2016.

\bibitem[Hermann et~al.(2015)Hermann, Kocisky, Grefenstette, Espeholt, Kay,
  Suleyman, and Blunsom]{hermann2015teaching}
Hermann, Karl~Moritz, Kocisky, Tomas, Grefenstette, Edward, Espeholt, Lasse,
  Kay, Will, Suleyman, Mustafa, and Blunsom, Phil.
\newblock Teaching machines to read and comprehend.
\newblock In \emph{Advances in Neural Information Processing Systems}, pp.\
  1693--1701, 2015.

\bibitem[Hochreiter \& Schmidhuber(1997)Hochreiter and
  Schmidhuber]{hochreiter1997long}
Hochreiter, Sepp and Schmidhuber, J{\"u}rgen.
\newblock Long short-term memory.
\newblock \emph{Neural computation}, 9\penalty0 (8):\penalty0 1735--1780, 1997.

\bibitem[Kadlec et~al.(2016)Kadlec, Schmid, Bajgar, and
  Kleindienst]{kadlec2016text}
Kadlec, Rudolf, Schmid, Martin, Bajgar, Ondrej, and Kleindienst, Jan.
\newblock Text understanding with the attention sum reader network.
\newblock \emph{arXiv preprint arXiv:1603.01547}, 2016.

\bibitem[Kiros et~al.(2015)Kiros, Zhu, Salakhutdinov, Zemel, Urtasun, Torralba,
  and Fidler]{kiros2015skip}
Kiros, Ryan, Zhu, Yukun, Salakhutdinov, Ruslan~R, Zemel, Richard, Urtasun,
  Raquel, Torralba, Antonio, and Fidler, Sanja.
\newblock Skip-thought vectors.
\newblock In \emph{Advances in neural information processing systems}, pp.\
  3294--3302, 2015.

\bibitem[Koutnik et~al.(2014)Koutnik, Greff, Gomez, and
  Schmidhuber]{koutnik2014clockwork}
Koutnik, Jan, Greff, Klaus, Gomez, Faustino, and Schmidhuber, Juergen.
\newblock A clockwork rnn.
\newblock \emph{arXiv preprint arXiv:1402.3511}, 2014.

\bibitem[Li et~al.(2016)Li, Tarlow, Brockschmidt, and Zemel]{li2015gated}
Li, Yujia, Tarlow, Daniel, Brockschmidt, Marc, and Zemel, Richard.
\newblock Gated graph sequence neural networks.
\newblock \emph{ICLR}, 2016.

\bibitem[Miller et~al.(2016)Miller, Fisch, Dodge, Karimi, Bordes, and
  Weston]{miller2016key}
Miller, Alexander, Fisch, Adam, Dodge, Jesse, Karimi, Amir-Hossein, Bordes,
  Antoine, and Weston, Jason.
\newblock Key-value memory networks for directly reading documents.
\newblock \emph{arXiv preprint arXiv:1606.03126}, 2016.

\bibitem[Miller et~al.(1990)Miller, Beckwith, Fellbaum, Gross, and
  Miller]{miller1990introduction}
Miller, George~A, Beckwith, Richard, Fellbaum, Christiane, Gross, Derek, and
  Miller, Katherine~J.
\newblock Introduction to wordnet: An on-line lexical database.
\newblock \emph{International journal of lexicography}, 3\penalty0
  (4):\penalty0 235--244, 1990.

\bibitem[Munkhdalai \& Yu(2016)Munkhdalai and Yu]{munkhdalai2016reasoning}
Munkhdalai, Tsendsuren and Yu, Hong.
\newblock Reasoning with memory augmented neural networks for language
  comprehension.
\newblock \emph{arXiv preprint arXiv:1610.06454}, 2016.

\bibitem[Onishi et~al.(2016)Onishi, Wang, Bansal, Gimpel, and
  McAllester]{onishi2016did}
Onishi, Takeshi, Wang, Hai, Bansal, Mohit, Gimpel, Kevin, and McAllester,
  David.
\newblock Who did what: A large-scale person-centered cloze dataset.
\newblock \emph{arXiv preprint arXiv:1608.05457}, 2016.

\bibitem[Oord et~al.(2016)Oord, Kalchbrenner, and Kavukcuoglu]{oord2016pixel}
Oord, Aaron van~den, Kalchbrenner, Nal, and Kavukcuoglu, Koray.
\newblock Pixel recurrent neural networks.
\newblock \emph{arXiv preprint arXiv:1601.06759}, 2016.

\bibitem[Paperno et~al.(2016)Paperno, Kruszewski, Lazaridou, Pham, Bernardi,
  Pezzelle, Baroni, Boleda, and Fern{\'a}ndez]{paperno2016lambada}
Paperno, Denis, Kruszewski, Germ{\'a}n, Lazaridou, Angeliki, Pham, Quan~Ngoc,
  Bernardi, Raffaella, Pezzelle, Sandro, Baroni, Marco, Boleda, Gemma, and
  Fern{\'a}ndez, Raquel.
\newblock The lambada dataset: Word prediction requiring a broad discourse
  context.
\newblock \emph{arXiv preprint arXiv:1606.06031}, 2016.

\bibitem[Rajpurkar et~al.(2016)Rajpurkar, Zhang, Lopyrev, and
  Liang]{rajpurkar2016squad}
Rajpurkar, Pranav, Zhang, Jian, Lopyrev, Konstantin, and Liang, Percy.
\newblock Squad: 100,000+ questions for machine comprehension of text.
\newblock \emph{arXiv preprint arXiv:1606.05250}, 2016.

\bibitem[Scarselli et~al.(2009)Scarselli, Gori, Tsoi, Hagenbuchner, and
  Monfardini]{scarselli2009graph}
Scarselli, Franco, Gori, Marco, Tsoi, Ah~Chung, Hagenbuchner, Markus, and
  Monfardini, Gabriele.
\newblock The graph neural network model.
\newblock \emph{IEEE Transactions on Neural Networks}, 20\penalty0
  (1):\penalty0 61--80, 2009.

\bibitem[Seo et~al.(2017{\natexlab{a}})Seo, Kembhavi, Farhadi, and
  Hajishirzi]{seo2016bidirectional}
Seo, Minjoon, Kembhavi, Aniruddha, Farhadi, Ali, and Hajishirzi, Hannaneh.
\newblock Bidirectional attention flow for machine comprehension.
\newblock \emph{ICLR}, 2017{\natexlab{a}}.

\bibitem[Seo et~al.(2017{\natexlab{b}})Seo, Min, Farhadi, and
  Hajishirzi]{seo2016query}
Seo, Minjoon, Min, Sewon, Farhadi, Ali, and Hajishirzi, Hannaneh.
\newblock Query-reduction networks for question answering.
\newblock \emph{ICLR}, 2017{\natexlab{b}}.

\bibitem[Shuai et~al.(2016)Shuai, Zuo, Wang, and Wang]{shuai2016dag}
Shuai, Bing, Zuo, Zhen, Wang, Bing, and Wang, Gang.
\newblock Dag-recurrent neural networks for scene labeling.
\newblock In \emph{Proceedings of the IEEE Conference on Computer Vision and
  Pattern Recognition}, pp.\  3620--3629, 2016.

\bibitem[Sukhbaatar et~al.(2015)Sukhbaatar, Weston, Fergus,
  et~al.]{sukhbaatar2015end}
Sukhbaatar, Sainbayar, Weston, Jason, Fergus, Rob, et~al.
\newblock End-to-end memory networks.
\newblock In \emph{Advances in neural information processing systems}, pp.\
  2440--2448, 2015.

\bibitem[Sutskever et~al.(2014)Sutskever, Vinyals, and
  Le]{sutskever2014sequence}
Sutskever, Ilya, Vinyals, Oriol, and Le, Quoc~V.
\newblock Sequence to sequence learning with neural networks.
\newblock In \emph{Advances in neural information processing systems}, pp.\
  3104--3112, 2014.

\bibitem[Tai et~al.(2015)Tai, Socher, and Manning]{tai2015improved}
Tai, Kai~Sheng, Socher, Richard, and Manning, Christopher~D.
\newblock Improved semantic representations from tree-structured long
  short-term memory networks.
\newblock \emph{arXiv preprint arXiv:1503.00075}, 2015.

\bibitem[Wang et~al.(2017)Wang, Onishi, Gimpel, and
  McAllester]{wang2017emergent}
Wang, Hai, Onishi, Takeshi, Gimpel, Kevin, and McAllester, David.
\newblock Emergent logical structure in vector representations of neural
  readers.
\newblock \emph{Preprint}, 2017.

\bibitem[Weston et~al.(2014)Weston, Chopra, and Bordes]{weston2014memory}
Weston, Jason, Chopra, Sumit, and Bordes, Antoine.
\newblock Memory networks.
\newblock \emph{arXiv preprint arXiv:1410.3916}, 2014.

\bibitem[Weston et~al.(2015)Weston, Bordes, Chopra, Rush, van Merri{\"e}nboer,
  Joulin, and Mikolov]{weston2015towards}
Weston, Jason, Bordes, Antoine, Chopra, Sumit, Rush, Alexander~M, van
  Merri{\"e}nboer, Bart, Joulin, Armand, and Mikolov, Tomas.
\newblock Towards ai-complete question answering: A set of prerequisite toy
  tasks.
\newblock \emph{arXiv preprint arXiv:1502.05698}, 2015.

\bibitem[Yang et~al.(2016)Yang, Blunsom, Dyer, and Ling]{yang2016reference}
Yang, Zichao, Blunsom, Phil, Dyer, Chris, and Ling, Wang.
\newblock Reference-aware language models.
\newblock \emph{arXiv preprint arXiv:1611.01628}, 2016.

\end{thebibliography}
\bibliographystyle{icml2017}

\end{document}